\documentclass{article}
\usepackage{spconf,amsmath,graphicx}
\usepackage{float}

\title{Multi-task Learning for Personal Health Mention Detection on Social Media}
\name{Olanrewaju Tahir Aduragba, Jialin Yu and Alexandra I. Cristea \thanks{Thanks to XYZ agency for funding.}}
\address{Author Affiliation(s)}
\address{Durham University\\
	Department of Computer Science\\
	Upper Mountjoy Campus, Stockton Road, Durham DH1 3LE, UK }
\begin{document}
%\ninept
%
\maketitle
\begin{abstract}
% The abstract should appear at the top of the left-hand column of text, about
% 0.5 inch (12 mm) below the title area and no more than 3.125 inches (80 mm) in
% length.  Leave a 0.5 inch (12 mm) space between the end of the abstract and the
% beginning of the main text.  The abstract should contain about 100 to 150
% words, and should be identical to the abstract text submitted electronically
% along with the paper cover sheet.  All manuscripts must be in English, printed
% in black ink.
Detecting personal health mentions on social media is essential to complement existing health surveillance systems. However, annotating data for detecting health mentions at a large scale is a challenging task. This research employs a multitask learning framework to leverage available annotated data from a related task to improve the performance on the main task to detect personal health experiences mentioned in social media texts. Specifically, we focus on incorporating emotional information into our target task by using emotion detection as an auxiliary task. Our approach significantly improves a wide range of personal health mention detection tasks compared to a strong state-of-the-art baseline.
\end{abstract}
\section{Introduction}
\label{sec:intro}
Social media platforms such as Twitter and Facebook have been useful in detecting signals for public health events. The real-time nature of these social data has made them useful for disease surveillance. The limitation of using traditional data sources for disease surveillance has created opportunities for disease using user-generated content from social media. However, the challenge with using these data sources is the large volume, the rate at which they are generated, the unstructured nature of the data, and potential biases. Natural language processing (NLP) techniques have been applied to social media data to support disease surveillance. NLP research in recent years has successfully represented social media texts to perform public health related downstream tasks. Some of these applications include monitoring the spread of disease outbreaks \cite{serban2019real} %public sentiment towards vaccination \cite{monselise2021topics} 
and detecting adverse drug reactions \cite{aduragba2020sentence}.

%, monitoring and supporting humanitarian emergencies \cite{yu2022multi} 
% and detecting adverse drug reactions \cite{aduragba2020sentence}.

To approach the problem of detecting health mention posts on social media, we explore the relationship between self-reports of personal health experiences and emotional expression. Health mentions are expected to trigger an emotion in the account poster. For example, someone reporting a diagnosis of a disease is likely to express emotions such as sadness, and fear, while someone who has recovered from an illness is likely to express emotions such as joy or happiness, even if they would be of similar health status, objectively; furthermore,  a post raising awareness about that particular disease might be neutral in terms of emotion. 

% As a result, 

Previous work on personal health mention detection has mostly considered the task on its own \cite{karisani2018did} or together with figurative usage detection \cite{iyer2019figurative}. In this paper, we leverage the emotions expressed in health mentions in a multi-task setting to improve the performance on the primary task: personal health mention detection.

% These guidelines include complete descriptions of the fonts, spacing, and
% related information for producing your proceedings manuscripts. Please follow
% them and if you have any questions, direct them to Conference Management
% Services, Inc.: Phone +1-979-846-6800 or email
% to \\\texttt{icip2022@cmsworkshops.com}.

\section{Related Work}
\label{sec:related_work}

% All printed material, including text, illustrations, and charts, must be kept
% within a print area of 7 inches (178 mm) wide by 9 inches (229 mm) high. Do
% not write or print anything outside the print area. The top margin must be 1
% inch (25 mm), except for the title page, and the left margin must be 0.75 inch
% (19 mm).  All {\it text} must be in a two-column format. Columns are to be 3.39
% inches (86 mm) wide, with a 0.24 inch (6 mm) space between them. Text must be
% fully justified.
\subsection{Personal Health Mention Detection}

Several works have proposed methods to detect personal health experiences on social media, utilising state-of-the-art techniques. Karisani and Agichtein \cite{karisani2018did} proposed a simple method (\emph{WESPAD} - Word Embedding Space Partitioning and Distortion) that combines lexical, syntactic, word embedding-based, and context-based features. Their approach aims to address the problems of sparsity and imbalanced training data for personal health mention detection. Their model learns to distort the word embedding space to more effectively distinguish cases of actual health mentions from the rest and partition the word embedding space to more effectively generalise from a small number of training examples.

Jing et al. \cite{jiang2018identifying} experimented with Long Short-Term Memory Networks (LSTM) to detect whether or not a tweet mentions a personal health experience. They applied generic pre-processing steps to the tweets before representing the individual tokens with pre-trained non-contextual word representations. Their approach outperforms conventional methods such as support vector machine (SVM), k-nearest neighbours algorithm (kNN), and decision tree models on the same task.

To address the problem of figurative use in disease and symptom terms in health mention task, Iyer et al. \cite{iyer2019figurative} jointly modelled figurative usage and personal health mention detection. They proposed a pipeline-based and feature augmentation-based approach to combine figurative usage detection with personal health mention detection. The feature augmentation-based approach performed best and used linguistic features and features extracted from unsupervised idiom detection. Both features are then concatenated and passed through a convolutional layer. 

Biddle et al. \cite{biddle2020leveraging} leveraged word-level sentiment distributions in addition to capturing figurative use of disease or symptom words to enhance performance on the health mention task. They showed that contextual word representations better classify health and figurative mentions than non-contextual word representations. Hence, they used contextual language models to generate word representations and used them in conjunction with sentiment distributions generated with a mixture of lexicon and neural methods. %Along the same line, Naseem et al. \cite{naseem2022identification} demonstrated that incorporating user behavioural information, including changes in emotions, thinking, or behaviour, can improve health mention tasks. They also consider literal word usage of disease or symptom terms in health mentions. 

In contrast to the above works, we propose incorporating emotional information through data-rich emotion detection tasks. We explore a multi-task learning approach where our primary task, personal health mention detection is learnt jointly with an auxiliary task, emotion detection.

\subsection{Emotion Detection}
There is a significant amount of research on emotion analysis on social media \cite{abdul2017emonet}. Researchers have explored several domains, ranging from generic \cite{SemEval2018Task1} %news \cite{oberlander2020goodnewseveryone} 
and natural disasters \cite{desai2020detecting}. Some of the work in the health domain explored detecting fine-grained emotions to track the emotional pulse of Twitter users in London, before and during the pandemic \cite{aduragba2021detecting}. Similarly, \cite{sosea2020canceremo} comprehensively analysed eight fine-grained emotions in an online health community and developed deep learning models to automatically detect them. They also vary in taxonomy, with the majority of existing public datasets including the 6 basic emotions categories (anger, disgust, fear, joy, sadness, and surprise) proposed by Ekman \cite{ortony1990s} and its extensions.

\section{Methodology}
\label{sec:methodology}

\begin{figure}
\centering
\includegraphics[width=0.4\textwidth]{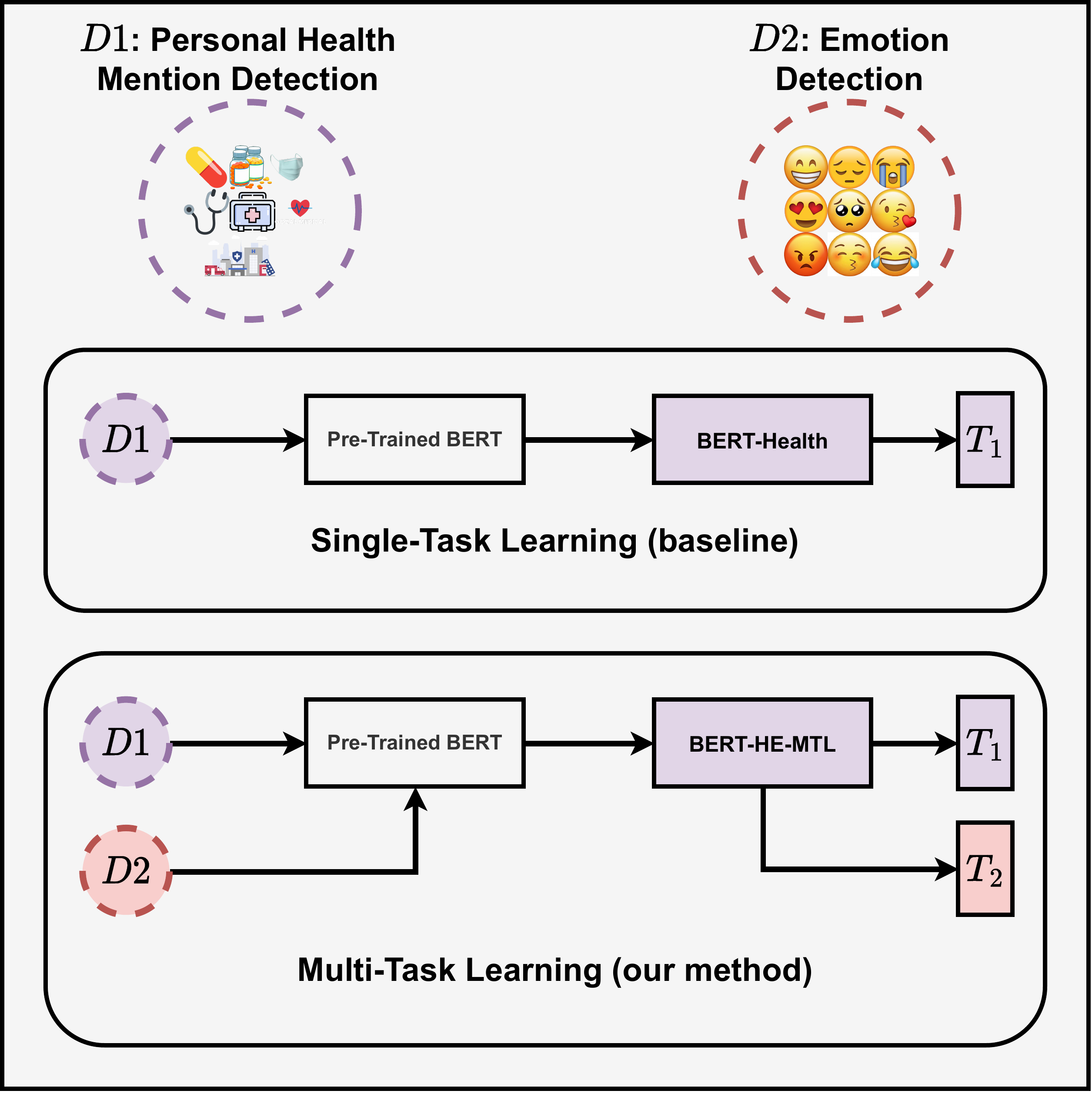}
\caption{Multitask learning framework to incorporate emotions in our main task of personal health mention, demonstrated with BERT model. The BERT model can be interchangeable with ALBERT and RoBERTa.}
\label{fig:framework}
\end{figure}

We experiment by integrating two sub-tasks of personal health mention detection and emotion detection in a multitask learning setting. This section gives an overview of our multitask learning framework and the parameter sharing scheme (Multitask Learning), with a graphical illustration of our framework in Fig. \ref{fig:framework}. We present a comparison baseline framework (Single Task Learning) for personal health mention detection, build based on a state-of-the-art model, BERT \cite{devlin2018bert}.

%with two diverse approaches to incorporate emotional information into the task of detecting personal health events on social media. These are Single Task Learning (STL) and Multitask Learning (MTL).

\subsection{Single Task Learning}
% Pre-trained language models have gained popularity in recent years because they learn useful knowledge that can be transferred to a downstream task. 
For single task learning, we treat the personal health mention detection task independently. We leverage the power of a pre-trained encoder to learn task-specific features from an input text. Large pre-trained language models such as BERT \cite{devlin2018bert} are especially appealing to this task because they adequately capture contextual information. %Specifically, we use the base version of BERT \cite{devlin2018bert}. 
Specifically, we use three different BERT-based models including BERT-base \cite{devlin2018bert}, Albert-base \cite{lan2019albert} and RoBERTa-base \cite{liu2019roberta}.

Given an input text, the BERT WordPiece tokenizer adds two special tokens \emph{[CLS]} and \emph{[SEP]}, at the beginning and the end respectively. Then, the tokenized text is passed into the BERT model to yield a sequence of vector hidden states $H = [h_{[CLS]}, h_1, h_2, ...,h_n, h_{[SEP]}]$. We consider the hidden vector $h_{[CLS]}$ from the last hidden layer to represent the aggregate representation of the text. Finally, we pass this representation into a single dense layer to predict the label.

% we fine-tune respective health mention datasets on a pre-trained language model.  Given an input and use the sentence representations in the $[CLS]$ token then feed into a linear layer to classify the final label. We use the single task learning models as our baseline. 

\subsection{Multitask Learning}
In our multitask learning framework (see Fig. \ref{fig:framework}), we explore the use of two optimization objectives: one for detecting personal health mentions, which is the primary task, and the other for detecting emotion in texts, as the auxiliary task. We hypothesise that the personal health mention task and emotion detection tasks are closely related and our primary task can benefit from additional sources of information from the training signals of related tasks. By sharing representation between the primary and auxiliary tasks, it will help our model generalise better on our primary task thereby improving performance. 

%To verify our hypothesis that personal health mention detection can benefit from an auxiliary task such as emotion detection, In our multitask learning setup we experiment with two setups. both soft parameter sharing and hard sharing, and a cross-stich network with . Soft-parameter sharing offers a way to   Specifically, we

\textbf{Parameter-sharing scheme}
There are two frequently used approaches to share parameters between tasks in a multitask learning setting: soft-parameter sharing and hard-parameter sharing \cite{ruder2017overview}. In multitask learning settings that employ hard parameter sharing \cite{caruana1997multitask}, both the primary and auxiliary tasks share a single encoder and the parameters are updated by both tasks while the shared encoder is followed by task-specific output layers. On the contrary, in soft parameter sharing, each task has task-specific encoders with their own parameters, and the distance between the parameters is regularised using a regularisation constraint to effectively share the parameters between tasks \cite{crawshaw2020multi}. 

As in \cite{liu2019multi}, we follow the approach used to train multiple tasks simultaneously by building task-specific layers on top of a pre-trained encoder. The pre-trained encoder shares its parameters across all tasks. Emotion expressions have been shown to be associated with social media utterances about personal health experiences \cite{metwally2017using}; hence in this paper, we adopt hard-parameter sharing to leverage emotional information to improve personal health mention detection performance. When the primary task and auxiliary task are closely related, hard-parameter sharing can be effective in improving performance on the primary task \cite{augenstein-etal-2018-multi}. 

% health care

% . For instance, it was discovered that, on average, tweets concerning colonoscopies expressed more negative sentiments than other topics \cite{metwally2017using}.

Given our primary task ($t_1$) and auxiliary task ($t_2$) and their corresponding data $D = {(X_1, Y_1), (X_2, Y_2)}$, where $(X_i, Y_i)$ is the training dataset for corresponding task $t_i$, the $i^{th}$ task is defined as follows:

\begin{equation}
    T^i(x, \theta) = \tau^i(\psi(x, \theta_{\psi}), \theta_i)
\end{equation}

 $\tau^i$ is the output of the $i^{th}$ task-specific module (with parameter $\theta_i \subset \theta$) and $\psi$ is the shared encoder (with parameter $\theta_{\psi} \subset \theta$). The goal of our multitask learning setup is to minimise the sum of the individual tasks losses:

\begin{equation}
    L = \lambda l_{1} + (1 - \lambda) l_{2}
\end{equation}

where $l_{1}$ and $l_{2}$ are the loss functions for the primary and auxiliary tasks respectively. $\lambda$ is a  hyperparameter that determines the weight that controls the importance we place on each task. In practice, we down-weight the loss of the auxiliary task to reduce the contribution of the task in updating the parameters. The loss function $l_i$ is as defined in equation \ref{loss_fn}.

\begin{equation}
    l_i = -\sum_{i=1}^Ny_i \cdot \log\hat{y}_i
\label{loss_fn}
\end{equation}

%We leverage the power of BERT as our shared encoder.
%We formulate our multitask learning as follows: Given labelled datasets $D = {(X_1, Y_1), (X_2,Y_2),...,(X_T, Y_T)}$ for $T$ tasks, where $t_i$ could be either our primary task (PHM) or auxiliary task (emotion detection). Our MTL model consists of a shared encoder that shares its parameters across tasks and it has task-specific layers with fully connected layer followed by softmax for classification. The softmax layers output probability distribution for each task according the following equations:

%A common way to train on multiple tasks is to select a batch of training examples from each task, cycling through them in fixed order. However, this may hurt the performance on the primary task. The imbalance could lead to over-fitting if the dataset from the primary task is smaller or under-training is the dataset of the primary task is larger.  To distinguish main tasks from auxiliary tasks the loss of the auxiliary task is down-weighted by a factor \lambda such that it comprises 10\% of the loss of the main task. λ is initialised with 1/10 and computed dynamically as training progresses.

\section{Experimental Setup}
\label{sec:exp_setup}

% To achieve the best rendering both in printed proceedings and electronic proceedings, we
% strongly encourage you to use Times-Roman font.  In addition, this will give
% the proceedings a more uniform look.  Use a font that is no smaller than nine
% point type throughout the paper, including figure captions.

% In nine point type font, capital letters are 2 mm high.  {\bf If you use the
% smallest point size, there should be no more than 3.2 lines/cm (8 lines/inch)
% vertically.}  This is a minimum spacing; 2.75 lines/cm (7 lines/inch) will make
% the paper much more readable.  Larger type sizes require correspondingly larger
% vertical spacing.  Please do not double-space your paper.  TrueType or
% Postscript Type 1 fonts are preferred.

% The first paragraph in each section should not be indented, but all the
% following paragraphs within the section should be indented as these paragraphs
% demonstrate.

\subsection{Data}
\begin{table}
\resizebox{.80\columnwidth}{!}{
\centering
\begin{tabular}{lcc}
\hline
\textbf{Dataset} & \textbf{Labels} & \textbf{Size}\\ 
\hline
% FLU2013 &Awareness& 2,622 \\
% &Self-report&\\
PHM2017 & Non-health & 4,987 \\
&Awareness&\\
&Other-mention&\\
&Self-mention&\\
\hline
HMC2019 & Health mention & 14,051 \\ %2
& Other mention &  \\ % 1
& Figurative mention &  \\ % 0
\hline
SELF2020 &No self-disclosure& 6,550\\
&Possible self-disclosure&\\
&Clear self-disclosure&\\
\hline
ILL2021 &Negative& 22,660\\
&Positive&\\

\hline
\end{tabular}}
\caption{Summary of health mention datasets}
\label{tab:summary_datasets}
\end{table}
To study the general applicability of our approach, we explore a variety of datasets from Twitter that are related to personal health mentions. A summary of all the datasets is provided in Table \ref{tab:summary_datasets}.

% \textbf{FLU2013}: this dataset was presented by \cite{lamb2013separating} and it focused on distinguishing actual infection reports of the disease from awareness of the disease. The original datasets consist of 4,516 tweets manually annotated for Flu awareness or Flu report. Due to Twitter data sharing policy, only the tweet IDs were released publicly. Hence, only 2,622 tweets were available to download as at Fall 2019.

\textbf{PHM2017}: This dataset focuses on more than one disease and condition was constructed by \cite{karisani2018did}. In the corpus, they collected English tweets related to Alzheimer’s disease, heart attack, Parkinson’s disease, cancer, depression, and stroke and manually annotated them in terms of \emph{self-mention}, \emph{other-mention}, \emph{awareness} and \emph{non-health}. At the time this research was conducted, only 4,987 tweets were available to download.

\textbf{HMC2019}: This dataset was introduced by \cite{biddle2020leveraging}. They focused on the same diseases and conditions as in PHM2017 \cite{karisani2018did} and extended with four additional conditions: cough, fever, headache, and migraine. Since disease words might be figuratively used on social media, they consider the figurative mentions of such words in their annotation. Their dataset was manually annotated in terms of \emph{figurative mention}, \emph{other mention} and \emph{health mention}. At the time this research was conducted, only 14,051 tweets were available to download. %At the time of downloading this dataset, only x tweets were available.

\textbf{SELF2020}: SELF2020 consists of health-related posts covering a range of health issues collected from online health forums on patient.info and social media platforms (Facebook, Reddit, and Twitter) \cite{valizadeh2021identifying}. The dataset aims to create a benchmark for a novel task of identifying medical self-disclosures. %Specifically, the task aims at detecting if a post discloses symptoms, diagnoses, or other information related to health issues. 
The dataset is annotated with \emph{no self-disclosure, possible self-disclosure} and \emph{clear self-disclosure}. SELF2020 differs from the other datasets in terms of content. While the other datasets are mainly filtered based on specific health conditions or diseases, SELF2020 is randomly sampled to prevent focus on disease-specific characteristics of health mentions. Since the majority (88.1\%) of the posts are from patient.info, the dataset contains phrases and sentences that are mostly longer than the Twitter-based datasets. %To be consistent with the other datasets, we only train with data labelled with \emph{no self-disclosure} and \emph{clear self-disclosure}.

\textbf{ILL2021}: The ILL2021 dataset is an illness report dataset related to three different health conditions: Parkinson's disease, cancer and diabetes. \cite{karisani2021contextual}. The dataset is annotated for detecting if a tweet mentions the health condition and contain a health report.

\textbf{GoEmotions}  GoEmotions \cite{demszky2020goemotions} is a benchmark emotion dataset originally annotated with 27 diverse emotions and neutral. The dataset contains 58k Reddit comments. The authors further group the labels into 6 Ekman emotion groups and neutral. This is the variant we use for our experiments,  where $emotions$ = \{anger, disgust, fear, joy, sadness, surprise, and neutral\}. We incorporate emotional knowledge from related tasks of emotion detection to classify health mentions using this dataset.  The dataset has 43,410 samples in the training set, 5,426 samples in the validation set, and 5,427 samples in the test set.

Since the official splits for the health mention datasets were not provided, we performed an 80/10/10 split to create the train, validation, and test sets. For each run of our models, we use a different random seed to initialise the split.

% \begin{table}
% \centering
% \resizebox{.99\columnwidth}{!}{
% \begin{tabular}{lllll}
% \hline
% \textbf{Model} & \textbf{FLU2013} & \textbf{PHM2017} & \textbf{SELF2020} & \textbf{ILL2021}\\ 
% \hline
% BERT-STL & 0.8664 & 0.8233 & 0.7716 & 0.9114\\
% BERT-MTL &\textbf{0.8753} & \textbf{{}0.8323} & \textbf{0.7787} & \textbf{0.9227}\\
% \hline
% \end{tabular}}
% \caption{F1 macro score for the personal health mention detection task. $\uparrow$ denotes better results and $\downarrow$ denotes worse results.}
% % improvements of the highest score}
% \label{tab:results}
% \end{table}

\begin{table}
\centering
\resizebox{.99\columnwidth}{!}{
\begin{tabular}{l|c|c|c|c}
\hline
\textbf{Model}     & \textbf{PHM2017} & \textbf{HMC2019} & \textbf{SELF2020} & \textbf{ILL2021} \\\hline %& \textbf{FLU2013}
\multicolumn{5}{c}{\textbf{Single task models}}            \\ \hline
BERT           &  81.82     & 88.12 &   76.73       &  91.45       \\ %&  85.80 
ALBERT      &     67.07    & 79.29    &  59.61   &     82.79    \\ %&    82.38
RoBERTa      &   65.58   &  79.67 &  60.14    &    82.55     \\ \hline %&    82.64 
\multicolumn{5}{c}{\textbf{Multi-task models}}             \\\hline
BERT         &    82.37 $\uparrow$    & 88.28 $\uparrow$ &     77.44  $\uparrow$    &   91.81 $\uparrow$      \\ %&     85.86 $\uparrow$
ALBERT        &   69.93 $\uparrow$    &  79.00 $\downarrow$   &  60.86 $\uparrow$    &   83.55 $\uparrow$      \\ %&  81.92 \downarrow
RoBERTa    &    70.70 $\uparrow$   & 78.53 $\downarrow$ &   61.45 $\uparrow$      &   81.53 $\downarrow$     \\ \hline %&     82.24 \downarrow 

\end{tabular}}
\caption{F1 macro score for the personal health mention detection task. $\uparrow$ denotes better results and $\downarrow$ denotes worse results.}
\label{tab:results}
\end{table}
 
% \subsection{Baselines}
% The details of our models are as follows. We compare incorporating emotions in a single-task learning with BERT fine-tuned on emotion data with BERT infused with emotion information and the multi-task learning that uses emotion recognition as an auxiliary tasks and with the following baselines.

% \textbf{Pre-Trained Language Models} We fine-tune BERT \cite{devlin2018bert}, specifically bert-base-case.

\subsection{Hyperparameter Tuning}

We tune our model hyperparameters on the development dataset of our health mention datasets to find the best training configurations. Bayesian optimization was used to find the optimal value of hyperparameters. The range of hyperparameters is summarized as follows: batch size $\varepsilon$  \{32, 64, 128\}, learning rate $\varepsilon$  [1e-6, 1e-3], dropout $\varepsilon$  [0.0, 1.0], loss weight parameter $\lambda$ $\varepsilon$  [0.0, 1.0] for MTL experiments. 

\section{Results and Discussion}
\label{sec:result}
Table \ref{tab:results} shows the results for our multitask model. The table shows the mean averaged F$1$ scores for the test set, based on $5$ runs using different random seeds of 69556, 79719, 30010, 46921, and 25577. In general, multitask learning models (MTL) outperforms the single task learning with statistically significant ($p<0.05$, based on the Wilcoxon test) improvements over all three BERT, RoBERTa, and ALBERT single task learning (STL) models. 

From the table, for PHM2017 and SELF2020 datasets, we observe general improvements in performance with all three model architectures. While for HMC2019 and ILL2021 datasets, we observed overall improvements in BERT and ALBERT, but not in RoBERTa. The reason may be caused by the size of the training dataset and the nature of the problem being a simpler classification task. We observe that incorporating emotional information is more beneficial with moderate-size personal health mention detection datasets, as not all required information is embedded in the datasets. On the contrary, when the size of the dataset is large enough, incorporating emotional information benefits less for personal health mention detection tasks. The observations correspond to our intuition of human decision-making, when information is limited, related knowledge from another domain helps us understand the problem better; while enough information is provided, related knowledge can also be a distraction. Our observation in this paper shed light on future research venues for personal health mention detection tasks to study the effect of various sizes of personal health mention detection datasets when incorporating emotional information.

% training datasets with regards to various training datasets size.

% From the table, the most improvement is in the ILL2021 dataset. This could be because of the size of the training dataset and the nature of the problem being a simpler classification task. On the other hand, the least gain was seen in the SELF2020 dataset. The dataset annotators reported that \emph{possible self-disclosure} showed lower inter-annotator agreement thereby making the task more challenging.

% The overall lower performance gain in FLU2013, PHM2017 and SELF2020 could be due to the small size of the datasets. The model is not able to benefit from better generalization and shared learning.

% Major headings, for example, "1. Introduction", should appear in all capital
% letters, bold face if possible, centered in the column, with one blank line
% before, and one blank line after. Use a period (".") after the heading number,
% not a colon.

% \subsection{Subheadings}
% \label{ssec:subhead}

% Subheadings should appear in lower case (initial word capitalized) in
% boldface.  They should start at the left margin on a separate line.
 
% \subsubsection{Sub-subheadings}
% \label{sssec:subsubhead}

% Sub-subheadings, as in this paragraph, are discouraged. However, if you
% must use them, they should appear in lower case (initial word
% capitalized) and start at the left margin on a separate line, with paragraph
% text beginning on the following line.  They should be in italics.

\section{Conclusion and Future Work}
\label{sec:conclusion}
In this paper, we showed that, as per our initial hypothesis, health data discussion contains emotional content, which can be exploited when applying classification tasks on it. Here we offer solutions based on multi-task models. For future research, we consider different pipelines as well as single-task models to explore further in-depth the effect of adding emotion detection for health-related text processing tasks. Furthermore, we believe in exploring the most efficient data sampling strategy, as data imbalance is a common problem in multi-task learning.

\bibliographystyle{IEEEbib}
\bibliography{strings,refs}

\end{document}